\documentclass[10pt,twocolumn,letterpaper]{article}

\usepackage{cvpr}
\usepackage{times}
\usepackage{epsfig}
\usepackage{graphicx}
\usepackage{amsmath}
\usepackage{amssymb}
\usepackage{caption}
\usepackage{booktabs}
\usepackage{fancyvrb}
\usepackage[singlelinecheck=false]{caption}
\usepackage[usenames,dvipsnames]{xcolor}
\usepackage{upquote}
\usepackage{mathtools}
\usepackage{subfigure}
\usepackage{alltt}
\usepackage{xcolor}
\usepackage{listings}
\usepackage{dsfont}

\usepackage[pagebackref=true,breaklinks=true,letterpaper=true,colorlinks,bookmarks=false]{hyperref}

 \cvprfinalcopy 

\def\cvprPaperID{793} 

\ifcvprfinal\fi
\begin{document}

\title{Building a Large-scale Multimodal Knowledge Base System\\ for Answering Visual Queries}

\author{Yuke Zhu \qquad Ce Zhang
\qquad Christopher R\'{e}
\qquad Li Fei-Fei\\
Computer Science Department, Stanford University\\
{\tt\small \{yukez,czhang,chrismre,feifeili\}@cs.stanford.edu}
}

\maketitle

\begin{abstract}
The complexity of the visual world creates significant challenges for comprehensive visual understanding. 
In spite of recent successes in visual recognition, today's vision systems would still struggle to deal with visual queries that require a deeper reasoning.
We propose a knowledge base (KB) framework to handle an assortment of visual queries, without the need to train new classifiers for new tasks.
Building such a large-scale multimodal KB presents a major challenge of scalability. 
We cast a large-scale MRF into a KB representation, incorporating visual, textual and structured data, as well as their diverse relations.
We introduce a scalable knowledge base construction system that is capable of building a KB with half billion variables and millions of parameters in a few hours.
Our system achieves competitive results compared to purpose-built models on standard recognition and retrieval tasks, while exhibiting greater flexibility in answering richer visual queries.
\end{abstract}

\section{Introduction}

\begin{figure}
\centering
\includegraphics[width=1.\linewidth]{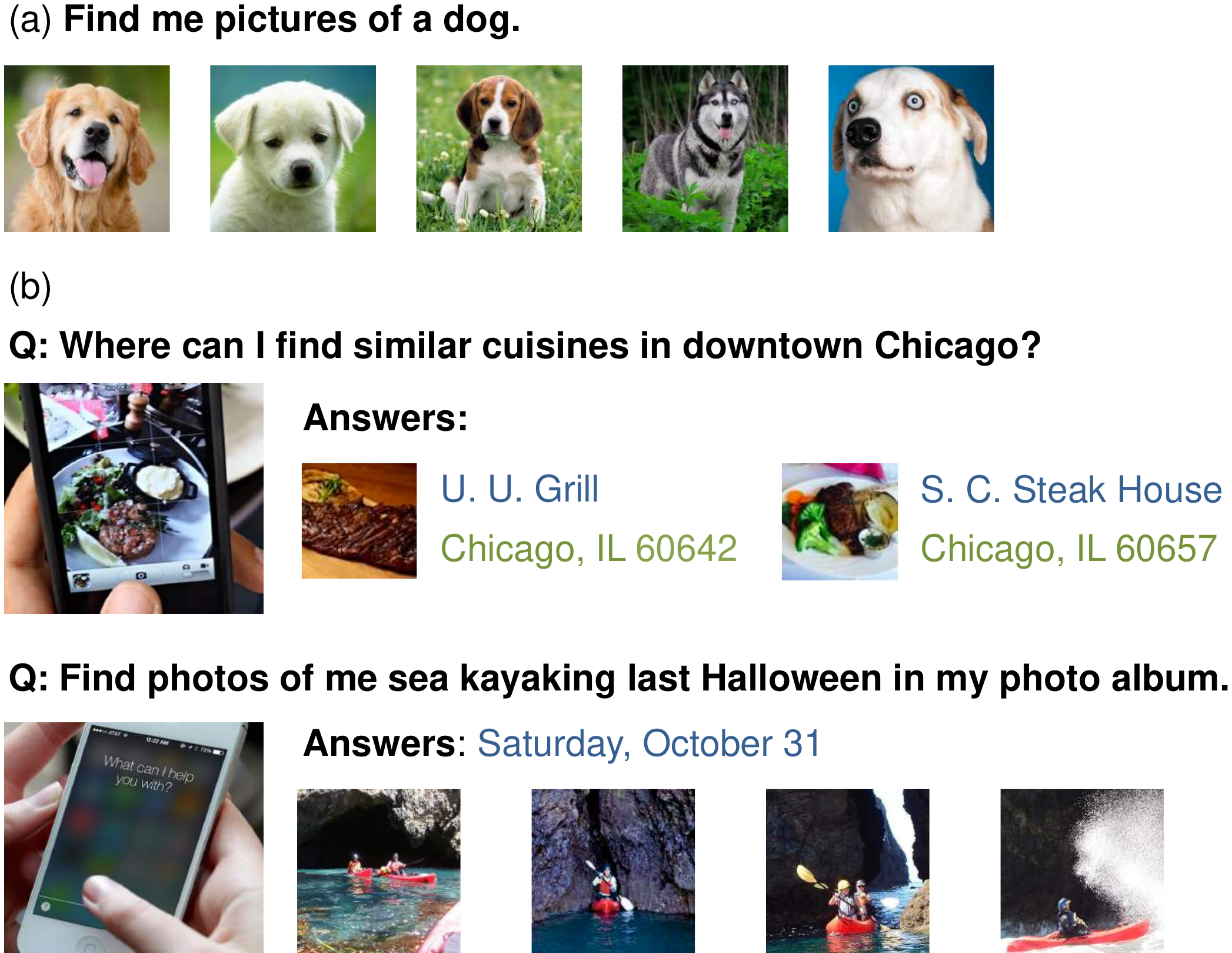}
\caption{\small{Although a classification-based method might be sufficient to find images of a dog in query (a). It would struggle for queries in real-world applications. To answer the queries in (b), we need to fuse visual information with metadata for joint reasoning. We propose a visual knowledge base framework to perform different types of visual tasks without training new classifiers. Our framework allows one to express this complex task with a single query.}}
\label{ex:one}
\vspace{-3mm}
\end{figure}

Type the following query in Google (i.e., a search engine) -- ``names of universities in Manhattan''. The returned list of answers is often sensible. But try this one -- ``names of universities with computer science PhD program in Manhattan''. The answers are far from satisfying. Both questions are perfectly clear to most humans, but current NLP-based algorithms still fail to perform well for more complex queries. In vision, we see a similar pattern. Much progress has been made in tasks such as classification and detection on single objects (e.g., Fig.~\ref{ex:one}(a)). But real-world vision applications might require more diverse and heterogeneous querying needs (e.g., Fig.~\ref{ex:one}(b)). The traditional classification-based methods would struggle in such tasks.

Towards the goal of scaling up the large-scale, diverse and heterogeneous visual querying tasks, a handful of recent papers~\cite{chen2013iccv,zhu2014eccv} have suggested to cast the visual recognition tasks into a framework that enables more heterogeneous reasoning and inference. A major benefit of doing so is to avoid training a new set of classifiers every time a new type of questions arises.  We approach this problem by building a large-scale multimodal \emph{knowledge base} (KB), where we answer visual queries (like the ones in Fig.~\ref{ex:one}(b)) by evaluating probabilistic KB queries.

A KB can often be viewed as a large-scale graph structure that connects different entities with their relations~\cite{Richardson06ml,zhu2009www}.
In NLP, some early promising results have been shown by encoding entity and relation information in text-based KBs, e.g., Freebase~\cite{bollacker2008sigmod} and IBM Waston's Jeopardy system~\cite{ferrucci2010}. In vision, there is now a small but growing amount of attention in building visual KBs.
In NEIL~\cite{chen2013iccv}, Chen et al. have shown the benefit of using contextual relations between scenes, objects and attributes to improve scene classification and object detection. However, its testing scenario is limited on recognition-based tasks; while it lacks a coherent inference model to extend to richer high-level tasks without training new classifiers. Zhu et al.~\cite{zhu2014eccv} have shown how to build a Markov Logic KB for affordance reasoning. However, their testing scenario is limited by its small data size and the discrete representation.
Our paper is particularly inspired by these two works~\cite{chen2013iccv,zhu2014eccv}, but focuses on addressing the following two key challenges.

First, \textbf{answering a variety of heterogeneous visual queries without re-training}. In real-world vision applications, the space of possible queries is huge (even infinite). It is impossible to retrain classifiers for every type of queries. Our system demonstrates its ability to perform reasoning and inference on an assortment of visual querying tasks, ranging from scene classification, image search to real-world application queries, without the need to train new classifiers for new tasks. 
We formalize answering these visual queries as computing the marginal probabilities of the joint probability model (Sec.~\ref{sec:qa-setup}). 
The key technique is to express visual queries in a logical form that can be answered from the visual KB in a principled inference method. 
We qualitatively evaluate our KB model in answering application queries like the ones in Fig.~\ref{ex:one}  (Sec.~\ref{sec:exp-qa}).
We then perform quantitative evaluations on the recognition tasks (Sec.~\ref{sec:scene-recognition-annotation}) and retrieval tasks (Sec.~\ref{sec:image-search-by-text}) respectively using the SUN dataset~\cite{xiao2010cvpr}. Our system achieves competitive results compared to the classification-based baseline models, while exhibiting greater flexibility in answering a variety of visual queries.

Second, \textbf{learning with large-scale multimodal data}. To build such a scalable KB, the model needs to perform joint learning and inference on a large amount of images, text and structured data, especially by using both discrete and continuous variables. Existing text-based KB representations~\cite{getoor2001springer,Richardson06ml,zhu2009www} fail to incorporate continuous visual features in a probabilistic framework, which hinders us from expressing richer multimodal data. In vision, MRFs have been widely used as a probabilistic framework to model joint distributions among multimodal variables.
We cast a MRF model into a KB representation to accommodate a mixture of discrete and continuous variables in a joint probability model. 
While MRFs have been widely used in a variety of vision tasks~\cite{Desai09iccv,kumar2003iccv,ladicky2010eccv,Torralba05nips}, applying them to a large-scale KB framework means that we need to conquer the challenge of scalable learning and inference. We build a scalable visual KB construction system by
leveraging database techniques, high-speed sampling~\cite{zhang2014pvldb} and first-order methods~\cite{niu2011nips}. We are able to build a KB with half billion variables and four million parameters, which is four orders of magnitude larger than Zhu et al.~\cite{zhu2014eccv} while using half of its training time.

\section{Previous Work}

\vspace{-1mm}
\paragraph*{Joint Models in Vision} 
A series of context models have leveraged MRFs in various vision tasks, such as
 image segmentation~\cite{he2014exemplar,ladicky2010eccv,mottaghi2014cvpr}, object recognition~\cite{Desai09iccv,kumar2003iccv}, object detection~\cite{Torralba05nips}, pose and activity recognition~\cite{yao2010cvpr} and other recognition tasks~\cite{Hoiem06cvpr,parikh2012pami}. Similarly, the family of And-Or graph models~\cite{tu2014joint,zhao2013cvpr} focus on parsing images and videos into a hierarchical structure. In this work, we use an MRF representation for joint learning and inference of our data, casting MRF models into modern KB systems. In particular, we address the scalability challenge of large-scale MRF learning with our knowledge base construction system.

\vspace{-4mm}
\paragraph{Learning with Vision and Language}
Previous work on joint learning with vision and language abounds~\cite{kong2014you,lin2015imagination,siddiquie2011cvpr,socher2013tacl,zitnick2013iccv}. Image and video captioning has recently become a popular task, where the goal is to generate a short text description for images and videos~\cite{chen2015cvpr,donahue2015cvpr,karpathy2015cvpr,lin2015generating,sutskever2014nips,vinyals2015cvpr,xu2015icml}.
It is followed by visual question answering~\cite{antol2015iccv,gao2015nips,malinowski2014multi,malinowski2015iccv,yu2015visualmadlibs}, which aims at answering natural language questions based on image content. Both captioning and question answering tasks perform on a single image and produce NLP outputs. Our system offers one single, coherent framework that can perform joint learning and inference on one or multiple images as well as metadata in textual and other forms.

\vspace{-4mm}
\paragraph*{Knowledge Bases}
Most KB work in the database and NLP communities focuses on organizing and retrieving only textual information in a structured representation \cite{bollacker2008sigmod,ferrucci2010,lenat1995acm,zhu2009www}. 
Although a few large-scale KBs~\cite{bollacker2008sigmod,dong2014kdd} have made attempts to incorporate visual information, they simply cache the visual contents and link them to text via hyperlinks.
In vision, a series of work has focused on extracting relational knowledge from visual data~\cite{chao2015cvpr,sadeghi2015viske,zitnick2013iccv}.
Chen et al.~\cite{chen2013iccv}, Divvala et al.~\cite{divvala2014cvpr} and Zhu et al.~\cite{zhu2014eccv} have recently proposed KB-based frameworks for visual recognition tasks. However, they all lack an inference framework to deal with more diverse types of visual queries.
PhotoRecall~\cite{photorecall_vsm2014} proposed a pre-defined knowledge structure to retrieve photos from text queries. In contrast, our system allows for new KB structures and offers the flexibility of answering richer types of queries.

\section{A Joint Probability Model: Casting a Large-Scale MRF into a KB System}
Our first task is to build a system that can efficiently learn a KB given a large amount of multimodal information, such as images, metadata, textual labels, and structured labels. Towards a real-world, large-scale system like this, the challenges are two-fold. First, our learning system must allow for a coherent probabilistic representation of both discrete and continuous variables to accommodate the heterogeneity of the data. Second, we need to develop an efficient but principled learning and inference method that is capable of large-scale computation. We address the first property in this section, and the second in Sec.~\ref{sec:kbc}. 

\subsection{The Knowledge Base System}
\label{sec:representation}
\begin{figure}
\centering
\includegraphics[width=.9\linewidth]{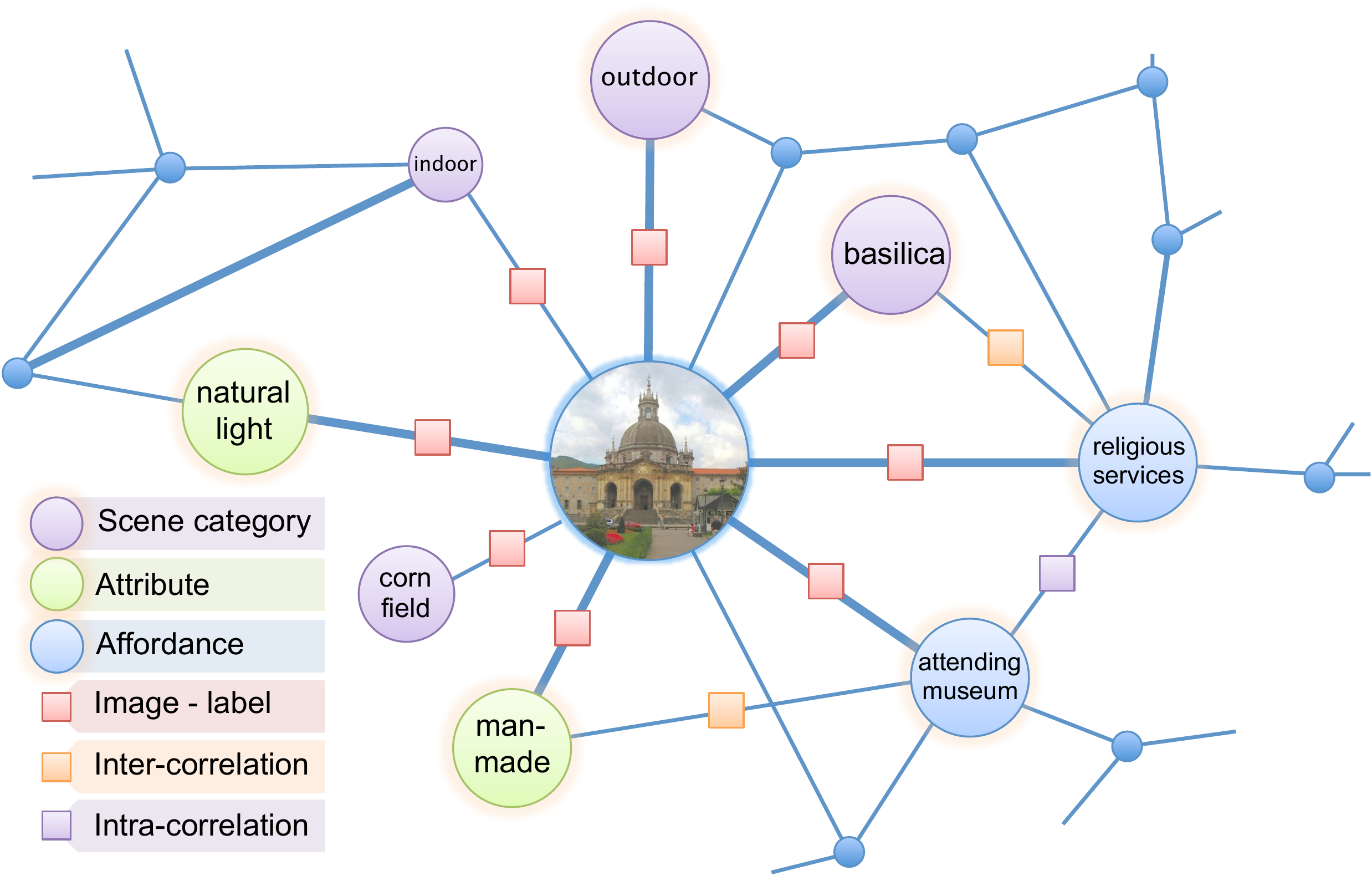}
\vspace{-2mm}
\caption{\small{\textbf{A graphical illustration of a visual knowledge base (KB).} A visual KB contains both visual entities (e.g., scene images) and textual entities (e.g., semantic labels) interconnected by various types of edges characterizing their relations. The nodes and edges correspond to the variables and factors respectively in the factor graph. The colors indicate different node (edge) types.}}
\label{fig:kb-vis}
\vspace{-2mm}
\end{figure}

A KB can be intuitively thought of as a graph of nodes connected by edges as in Fig.~\ref{fig:kb-vis}, 
where the nodes are called ``entities'' and the edges are called ``relations''. 
In vision, MRFs have been widely used to represent such graph structures~\cite{Hoiem06cvpr,mottaghi2014cvpr,parikh2012pami,Torralba05nips}. 
Thus, we cast an MRF model as the KB representation, where entities are represented by variables and relations by edges between variables.
This model provides an umbrella framework for answering visual queries, where we formalize query answering as evaluating marginals from the joint distribution (Sec.~\ref{sec:qa-setup}). 
In comparison to MLNs used in previous work~\cite{Richardson06ml,zhu2014eccv}, this representation is more generic, allowing us to accommodate continuous random variables and real-valued factors.
In practice, we use factor graphs~\cite{Kschischang01it,wick2010scalable}, a bipartite graph equivalence of an MRF. 
Factor graphs provide a simple graphical interpretation of the MRF model, resulting in ease of implementation for large-scale inference. 

A factor graph has two types of nodes: \emph{variables} and \emph{factors}. 
A possible world is a particular assignment to every variable, denoted by $I$. We define the probability of a possible world $I$ to be proportional to a log-linear combination of factors. We assign different weights to factors, expressing their relative influence on the probability. Formally, we define the {\em partition function} $Z$
of a possible world $I$ as
\begin{equation}
Z[I] = \exp\left(\sum_{i=1}^{m}w_if_i(I)\right)
\end{equation}
where $w_i$ is the weight of the $i$-th factor, $f_i(I)$ is the value of the $i$-th factor in possible world $I$, and $m$ is the total number of factors. The probability of a possible world is
\begin{equation}
\Pr[I;\mathbf{w}] = Z[I]\left(\sum_{I' \in \mathcal{I}}Z[I']\right)^{-1}
\label{eq:probability_world}
\end{equation}
where $\mathcal{I}$ is the set of all possible worlds, and $\mathbf{w}$ corresponds to the factor weights. In Fig.~\ref{fig:kb-vis}, each node corresponds to a variable; and each edge between nodes corresponds to a factor.  We define all the factors used in our KB in Sec.~\ref{sec:datasource}.

Having defined the structure of the factor graph KB, our learning objective is to find the optimal weight 
\begin{equation}
\mathbf{w}^{*}=\arg\min_{\mathbf{w}}-\sum_{I\in\mathcal{I}_{E}}\log\Pr[I;\mathbf{w}]+\lambda||\mathbf{w}||^2_2
\label{eq:learning_objective}
\end{equation}
where $\mathcal{I}_{E}$ is the set of possible worlds obtained from the training images and $\lambda$ is the regularization parameter. To optimize Eq.~\eqref{eq:learning_objective}, we need to compute the stochastic gradient $\frac{\partial\Pr[I|\mathbf{w}]}{\partial\mathbf{w}}$. It is usually intractable to compute the analytical gradients, as it involves the computation of an expectation over all possible words. We use the contrastive divergence scheme~\cite{hinton2002training} to estimate the log-likelihood gradients. The gradient of the weight $w_i$ of the $i$-th factor (omitting regularization) is approximated by:
\begin{equation}
\nabla w_i \approx  f_i(I') - f_i(I'')
\label{eq:approximate-gradient-partial}
\end{equation}
where $I'$ is a possible world sampled from the training data, and $I''$ is a possible world sampled under the distribution formed by the model (parameterized by $\mathbf{w}$). 
Gibbs sampling~\cite{chen2003continuous,he2004multiscale} is used as the transition operator of the Markov chain.
Intuitively, the first term in Eq.~\eqref{eq:approximate-gradient-partial} increases the probability of training data; and the second term decreases the probability of samples generated by the model. 
In-depth studies on the estimated gradients of Eq.~\eqref{eq:approximate-gradient-partial} can be found in the context of RBM training~\cite{carreira2005aistats,tielemanPcd2008}.
We show in Sec.~\ref{sec:kbc} that our system automatically creates a factor graph and learns the weights in a principled and scalable manner.

\subsection{Data Sources for the Knowledge Base}
\label{sec:datasource}
We now describe the entities and relations in our KB, and the data sources that we will use to populate the KB. For our purposes, SUN~\cite{xiao2010cvpr} is a particularly useful dataset because of a) its diverse set of images, and b) the availability of a large number of category and attribute labels.

\vspace{1mm}
\noindent
\textbf{Entities} can be thought of as descriptors of the images. In the factor graph depicted in Fig.~\ref{fig:kb-vis}, they are the nodes (variables) of the graph.

\vspace{1mm}
\emph{Images} -- are represented by their 4096-dimensional activations from the last fully-connected layer in a convolutional network \cite{zeiler2014eccv}. In total, there are 59,709 images from the SUN dataset~\cite{xiao2010cvpr}, where half are used for building the KB, and half for evaluation.

\vspace{1mm}
\emph{Scene category labels} -- indicate scene classes. In our experiments, we use 15 basic-level categories (e.g., workplace and transportation), and 298 fine-grained level categories (e.g., grotto and swamp) from SUN~\cite{xiao2010cvpr}.

\vspace{1mm}
\emph{Attribute labels} -- characterize visual properties (e.g., material, layouts, lighting, etc.) of a scene. We use the SUN Attribute Dataset~\cite{patterson2014ijcv}, which provides 102 attribute labels (e.g., glossy and warm).

\vspace{1mm}
\emph{Affordance labels} -- describe the functional properties of a scene, i.e., the actions that one can perform in a scene. We use a lexicon of 227 affordances (actions).\footnote{from the American Time Use Survey (ATUS)~\cite{shelley2005mlr} sponsored by the Bureau of Labor Statistics, which catalogs the actions in daily lives and represents United States census data} We conducted a large-scale online experiment to annotate the possibilities of the 227 actions for each scene category.  
We provide the list of affordances in Sec.~\ref{sec:affordance-list} in the supplementary material.
\vspace{1mm}

\vspace{1mm}
\noindent
\textbf{Relations} link entities (variables) to each other, as depicted by the squares on the edges in Fig.~\ref{fig:kb-vis}. The weights learned for the edges (factors) indicate the strength of the relations.
We introduce three types of relations in our model. 

\vspace{1mm}
\emph{Image\,-\,label} -- maps image features to semantic labels.

\emph{Intra-correlations} -- capture the co-occurrence between attribute-attribute and affordance-affordance pairs.

\emph{Inter-correlations} -- characterize correlations between two different types of labels (category\,-\,affordance, affordance\,-\,attribute, category\,-\,attribute and relations between categories from different levels). 
\vspace{2mm}

The entities and relations in the KB are mapped to variables and factors in the factor graph. 
We represent the image entities as continuous variables, and the label entities as discrete variables. 
Each image is associated with hundreds of attribute and affordance labels. Together, this amounts to a KB of millions of entities. Table~\ref{table:kb-statistics} summarizes some of the basic statistics of the KB that will be learned. This is two orders of magnitude larger than previous work~\cite{zhu2014eccv} regarding the number of entities and relations. 
The large size of our dataset presents a significant challenge of scalability.
In theory, an MRF can be arbitrarily large. However, its scalability is subject to the inefficiency of learning and inference. In addition, it is prohibitive to handcraft such a large-scale model from scratch. We, therefore, need a principled and scalable system for constructing the visual KB.


\begin{figure*}
\centering
\includegraphics[width=1.\linewidth]{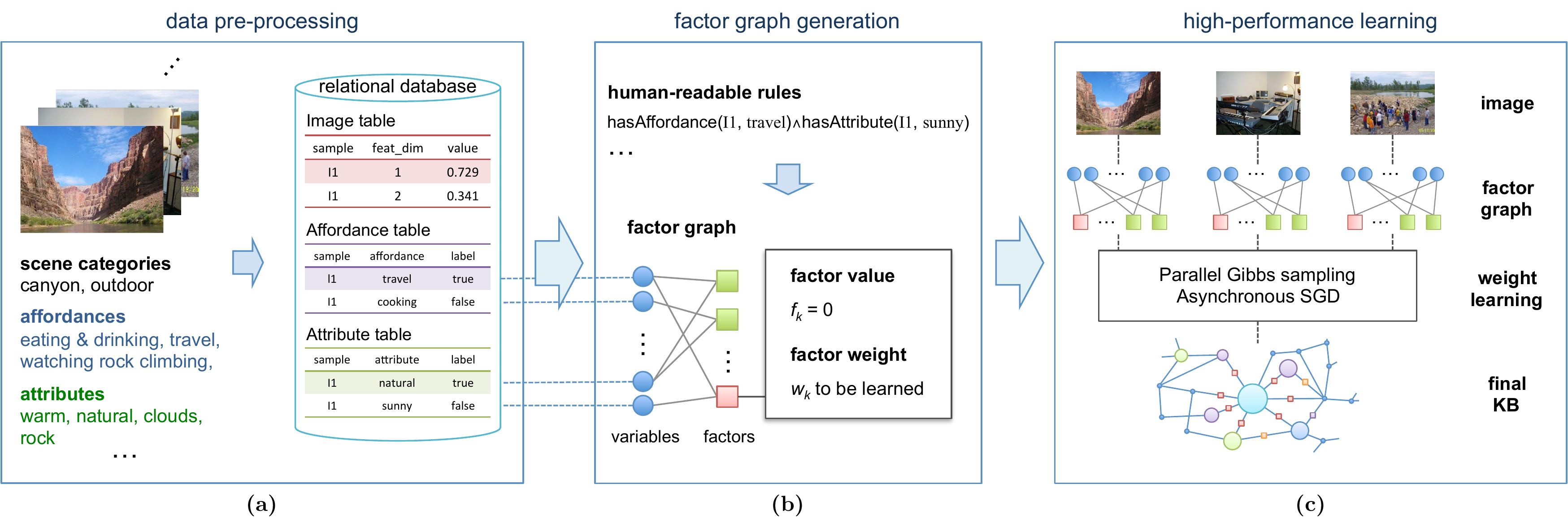}
\vspace{-7mm}
\caption{\small{\textbf{An overview of the knowledge base construction pipeline.} We first process the images and text, converting them into a structured representation. We write human-readable rules to define the KB structure. The system automatically creates a factor graph by parsing the rules. We then adopt a scalable Gibbs sampler to learn the weights in the factor graph.}}
\label{fig:kb-pipeline}
\vspace{-1mm}
\end{figure*}

\begin{table}[htp]
\centering
\vspace{-1mm}
\captionsetup{justification=centering}
\caption{\textbf{KB  Dataset Statistics}}
\vspace{-6mm}
\begin{center}
\begin{small}
\begin{tabular}{|l|c|c|}
\hline
& \textbf{Attributes} & \textbf{Affordances}\\
\hline
\hline
\textbf{Lexicon size} & 102 & 227\\
\textbf{\# Total labels} & $1.34\times 10^6$ & $1.36\times10^7$\\
\textbf{\# Positive labels} & $9.6\times 10^5$ & $1.23\times10^6$\\
\textbf{\# Positive / image} & 6.7 & 13.7\\
\hline
\end{tabular}
\end{small}
\end{center}
\label{table:kb-statistics}
\vspace{-7mm}
\end{table}%

\vspace{-1mm}
\section{Learning the Large-scale KB System}
\label{sec:kbc}
\vspace{-1mm}
Given our goal towards learning a real-world, large-scale MRF-based KB system, the biggest challenge we need to address here is efficient learning and inference. A number of recent advances have been made in the database community to shed light on how to build a large-scale KB~\cite{dong2014kdd,ferrucci2010,niu2011vldb}. Our framework follows closely that of Niu et al.~\cite{niu2011vldb}. In addition to that, we address the challenge of learning with multimodal data.
 Our KB system and the data will be made available to the public.

\subsection{Scalable Construction}
\label{sec:scalability}
There are three key steps to make the knowledge base construction (KBC) scalable: data pre-processing, factor graph generation and high-performance learning. 
Fig.~\ref{fig:kb-pipeline} offers an overview of the KBC process illustrating these three steps, which are indicated by the boxes.

\vspace{-4mm}
\paragraph*{Data Pre-processing}
The first step (the first box in Fig.~\ref{fig:kb-pipeline}) is to pre-process raw data into a structured representation, in particular, as tables in a relational database. Each database table stores the entities of the same type (e.g., the Affordance table in Fig.~\ref{fig:kb-pipeline}(a)).
It provides us access to database techniques such as SQL queries and parallel computing, important to achieve high scalability. 
We provide the database schema in Sec.~\ref{sec:kbc-supp} in the supplementary material.

\vspace{-4mm}
\paragraph*{Factor Graph Generation}
We represent the MRF model by a factor graph for the ease of implementation for scalable learning.
The factor graph is generated from the database tables (the second box in Fig.~\ref{fig:kb-pipeline}). 
Each row in the database tables corresponds to a variable in the factor graph. 
For each training image, we construct a factor graph, where the variables (blue circles in Fig.~\ref{fig:kb-pipeline}(b)) are linked to their values in the database (dashed lines between Fig.~\ref{fig:kb-pipeline}(a) and (b)). 
We then define the factors on these variables.
It is prohibitive to handcraft a large KB structure. Instead, we develop a declarative language that allows us to define the factors with a handful of human-readable rules.
This language is a simple but powerful extension to previous work like MLNs~\cite{Richardson06ml} and PRMs~\cite{getoor2001springer}, 
which enables us to specify relations between multimodal entities in logical conjunctions.
We show an example rule in Fig.~\ref{fig:kb-pipeline}(b). This rule describes co-occurrence between affordance label \texttt{travel} and attribute label \texttt{sunny} on image \texttt{I1}. It evaluates to 1 if both labels are true and 0 otherwise.
The KBC system parses this rule and creates a factor $f_k$ on these two variables. A weight $w_k$ is assigned to this factor and will be learned in the next step.
The system creates a small factor graph for each of the training images. There is no edge between these graphs; however, the same factors in the graphs share the same weight (illustrated by the red squares in Fig.~\ref{fig:kb-pipeline}(c)). The weight sharing scheme is also specified in the declarative language. We provide a detailed explanation of the declarative language and a complete list of rules in Sec.~\ref{sec:kbc-supp} in the supplementary material.

\vspace{-4mm}
\paragraph*{High-Performance Learning}
\label{sec:learning-and-inference}
Having defined the factor graph structures, our goal is to learn the factor weights efficiently.
We use the learning method in Sec.~\ref{sec:representation} to find the optimal factor weights.
We built a Gibbs sampler for high-performance learning and inference that is able to handle multimodal variables.
Our system performs scalable Gibbs sampling based on careful system design and speedup techniques. On the system side, we implemented the Hogwild! model~\cite{niu2011nips,zhang2014pvldb} which can run asynchronous stochastic gradient descent while still guaranteeing convergence. The system runs in parallel, allowing the sampler to achieve a high efficiency. On average, our Gibbs sampler processes $8.2\times 10^7$ variables per second. Finally this step produces a learned visual KB.

\subsection{Learning Efficiency}
\label{sec:learning_efficiency} 
\begin{figure}
\centering
\includegraphics[width=.7\linewidth]{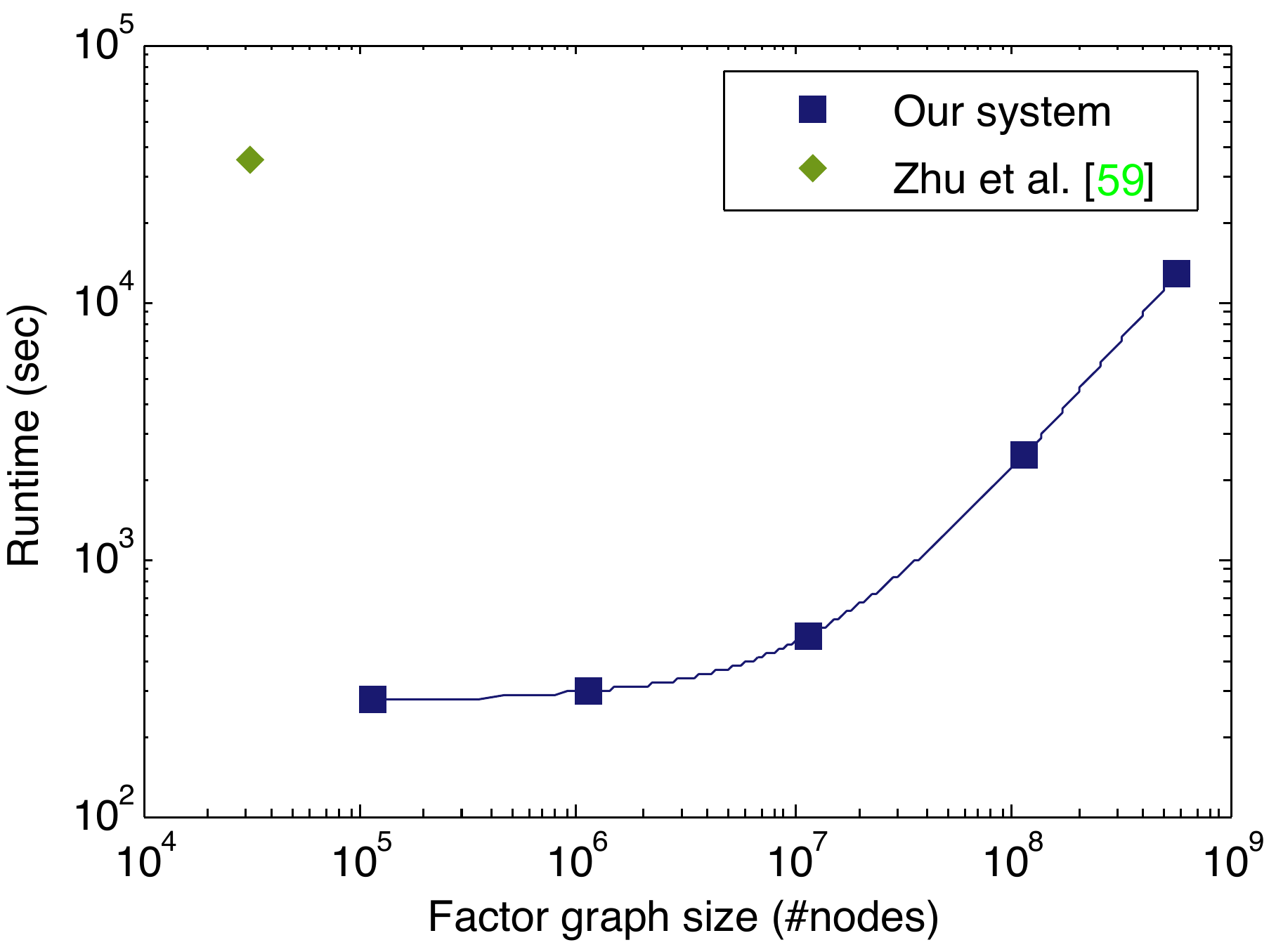}
\vspace{-2mm}
\caption{\small{\textbf{Efficiency of the knowledge base construction system.} The curve is plotted in log-log scale, where the $x$-axis is the number of nodes in the factor graph, and the $y$-axis is the runtime to construct the KB.}}
\label{fig:perf_curve}
\vspace{-2mm}
\end{figure}

The three steps (described in Sec.~\ref{sec:scalability}) together contribute to the high scalability of our KBC system. Table~\ref{table:fg-stats} shows that with this framework, we can build a KB four orders of magnitude larger regarding the number of variables and three orders of magnitude larger regarding model parameters compared to~\cite{zhu2014eccv} (using Alchemy MLNs~\cite{Richardson06ml}), in half of the time. Fig.~\ref{fig:perf_curve} further demonstrates that the learning time grows steadily as the KB size increases. 
The end-to-end construction finishes in 5.2 hours on the whole dataset (Sec.~\ref{sec:datasource}), indicating the potential to build larger-scale KBs in the future.

\begin{table}[htb]
\centering
\captionsetup{justification=centering}
\caption{\textbf{Statistics of the Visual KB Systems}}
\vspace{-5mm}
\begin{center}
\begin{tabular}{|cccc|}
\hline
& \textbf{variables} & \textbf{parameters} & \textbf{runtime} \\
\hline
\hline
\textbf{Zhu et al.} \cite{zhu2014eccv} & $3.15\times 10^4$ & $5.06\times 10^3$ & 10 hr\\
\textbf{Ours} & $5.76 \times 10^8$ & $4.19\times 10^6$ & 5.2 hr\\
\hline
\end{tabular}
\end{center}
\label{table:fg-stats}
\vspace{-5mm}
\end{table}

\section{Visual Query Setup}
\label{sec:qa-setup}
As we have mentioned in the introduction, one advantage of using a KB system is its ability to handle rich and diverse types of visual queries without training new classifiers. Moreover, this inference is done in one joint model without step-wise filtering, treating images and other metadata on an equal footing in learning and inference. From a user's perspective, the input to this system is a natural language question along with a set of one or more images. Similarly, the output is a mixture of images and text. 

In practice, the space of possible queries is huge. It would be prohibitive to map each natural language question to the corresponding inference task in an ad-hoc manner. 
One solution is to reformulate the questions in a formal language~\cite{berant2013semantic}, such as a probabilistic query language based on conjunctive queries~\cite{suciu2011morganclaypool}. This language allows us to express KB queries and to compute a ranked list of answers based on their marginal probabilities.
We briefly describe how this works by an example query that retrieves images of a sunny beach. This query is formed by a conjunction of two predicates (Boolean-valued functions) of \texttt{sceneCategory} and \texttt{hasAttribute}:
\begin{equation*}
\small
\texttt{sceneCategory}(i, \texttt{beach})\,\wedge\,\texttt{hasAttribute}(i, \texttt{sunny})
\end{equation*}
Given such a query, our task is to find all possible images $i$ where both predicates are true -- i.e.,  image $i$ comes from the scene category \texttt{beach} and has the attribute \texttt{sunny}. Following this example, more complex queries can be formed by joining several predicates together.\footnote{In this work, we manually annotate the conjunctive queries from natural language questions. The mapping from sentences to logical forms is a well-studied problem in NLP~\cite{berant2013semantic} and orthogonal to our system.}

Let $Q$ be a conjunctive query such as the one above. 
We compute a ranked list of answers (e.g., images of sunny beaches) based on their marginal probabilities. 
Formally, the marginal probability of a tuple $t$ (a list of variable assignments) being an answer to $Q$ is defined as:
\begin{equation}
\Pr[t\in Q]=\sum_{I\in\mathcal{I}}\mathds{1}_{t\in Q(I)}\cdot\Pr[I;\mathbf{w}]
\label{eq:answer_prob}
\end{equation}
where $\mathcal{I}$ and $\Pr[I;\mathbf{w}]$ are defined in Eq.~\eqref{eq:probability_world}, $\mathds{1}$ is the indicator function, and $Q(I)$ is the set of variable assignments in the possible world $I$ under which $Q$ evaluates to true. We use the same Gibbs sampler as in Sec.~\ref{sec:scalability} to estimate tuple marginals by sampling a collection of possible worlds and averaging the query values over these possible worlds. Each query evaluation produces a set of tuple-probability pairs $\{(t_1, p_1), (t_2, p_2), \ldots\}$, where we retrieve the top answers by sorting the pairs based on their probabilities in a descending order.

\section{Experiments}
\label{sec:exp}
Now that we have learned a large KB from multimodal data sources, and have established a probabilistic language to express visual queries, we can demonstrate how a KB can be useful in a number of querying tasks.
To demonstrate the utility of our KB, we perform several types of evaluations that involve vision tasks with multimodal answers including images, text and metadata.

\subsection{Answering Queries of Diverse Types}
\label{sec:exp-qa}
We start with a qualitative demonstration of using the KB to answer a wide variety of queries by performing joint inference on image appearance, as well as metadata like geolocations, timestamps, and business information.\footnote{These metadata are either acquired from existing databases or automatically scraped online. Detailed descriptions of the experimental setups and the conjunctive queries (Sec.~\ref{sec:qa-setup}) for Fig.~\ref{fig:crazy-queries} are provided in Sec.~\ref{sec:qa-app-supp} in the supplementary material.}
\begin{figure}[t]
\centering
\includegraphics[width=1.0\linewidth]{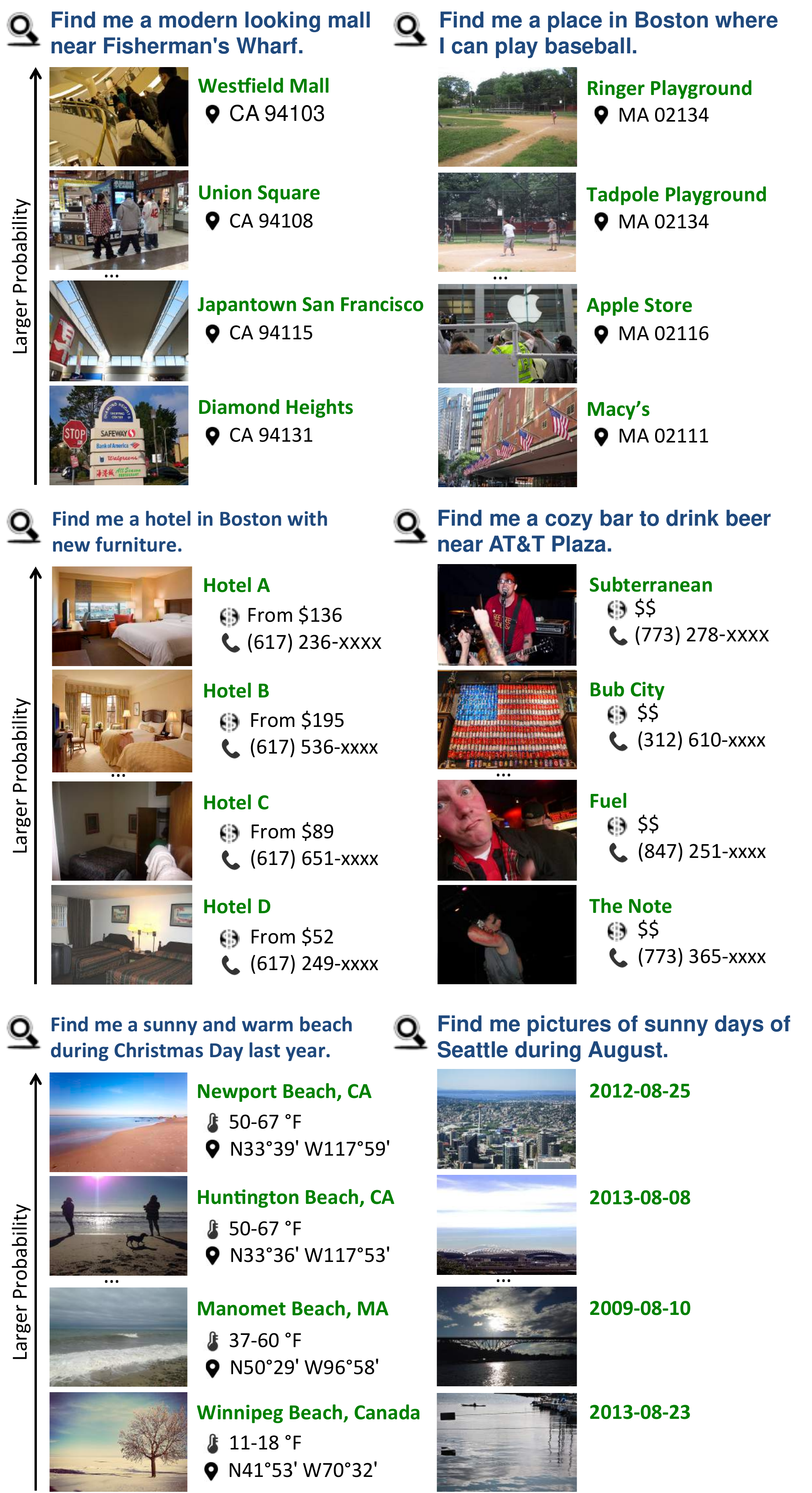}
\vspace{-6mm}
\caption{\small{\textbf{Proof-of-concept queries in a query answering application.} We incorporate external data to enrich our knowledge base, and demonstrate its flexibility in answering real-world queries.}}
\label{fig:crazy-queries}
\vspace{-5mm}
\end{figure}
Fig.~\ref{fig:crazy-queries} provides a few examples that depict the rich queries the system can handle. A user can ask the KB a question in natural language, such as ``find me a modern looking mall near Fisherman's Wharf.'' While the photos of the malls are not part of the training data in Sec.~\ref{sec:datasource}, our system is capable of linking the photo contents to other metadata, and is able to offer the names and locations of the shopping malls. Similarly in the second example ``find me a place in Boston where I can play baseball'', our system predicts the affordances from the appearances of the photos, and combines them with geolocation information to retrieve a list of places for playing baseball.
In Fig.~\ref{fig:crazy-queries}, the answers are shown in a ranked list by their marginal probabilities. Without a principled inference model, previous work such as NEILL\cite{chen2013iccv} and LEVAN~\cite{divvala2014cvpr} cannot produce such probabilisitic outputs.

\subsection{Single-Image Query Answering}
\label{sec:scene-recognition-annotation}
While our KB is designed for answering a wide range of queries, we can still evaluate how our system performs quantitatively in several standard visual recognition tasks without re-training. Based on the KB we have learned from data sources such as SUN (see Sec~\ref{sec:datasource}), we show two experiments for scene classification and affordance prediction. Both of these two tasks can be thought of as answering queries for a single image, where these queries can be expressed by a single predicate with the querying labels taken as random variables -- i.e., $\texttt{sceneCategory}(\texttt{img}, c)$ and $\texttt{hasAffordance}(\texttt{img}, a)$. Our system outperforms the state-of-the-art baseline methods for each of these tasks.

\begin{table}[t]
\captionsetup{justification=centering}
\caption{\textbf{Performance of Scene Classification (in mAcc)}}
\centering
\vspace{-2mm}
\begin{tabular}{|l|c|c|}
\hline
\textbf{Method} & \textbf{Basic level} & \textbf{Fine-grained}\\
\hline
\hline
CNN Fine-tuned \cite{zeiler2014eccv} & 89.1 & 67.5\\
Attribute-based model & 88.0 & 57.9\\
Attributes\,+\,Features & 90.2 & 69.6\\
\hline
\hline
KB - Affordances & 90.0 & 69.3\\
KB - Attributes & 90.7 & 69.6\\
KB -  Full & \textbf{91.2} & \textbf{69.8}\\
\hline
\end{tabular}
\label{table:scene-recogniton}
\vspace{-1mm}
\end{table}%

For both experiments, we use the data in Sec.~\ref{sec:datasource} for training and an evaluation set of 29,781 images from the same 298 categories of SUN~\cite{xiao2010cvpr} for testing.
We measure {\em scene classification} by mean accuracy (mAcc)  over classes~\cite{zhou2014places}. SUN~\cite{xiao2010cvpr} provides two ways of classification: basic-level (15 categories) and fine-grained (298 categories). Table~\ref{table:scene-recogniton} provides a summary of the results, comparing our full model (KB\,-\,Full) with a number of different settings and state-of-the-art models. We describe the models used in Table~\ref{table:scene-recogniton} as follow:

\begin{itemize}
\vspace{-1mm}
\item {\bf{CNN Fine-tuned}} We fine-tuned a CNN~\cite{zeiler2014eccv} on a subset of SUN397 dataset~\cite{xiao2010cvpr} of 107,754 images. We train $\ell_2$-logistic regression classifiers on the activations from the last fully-connected layer. We also use this as image features for all the other baselines.
\vspace{-1mm}
\item {\bf{Attribute-based model}} We predict the scene attributes and affordances from the CNN features, and use a binary vector of the predicted values as an intermediate feature. This is the strategy adopted by Zhu et al.~\cite{zhu2014eccv} to discretize visual data.
\vspace{-1mm}
\item	{\bf{Attributes\,+\,Features}} We concatenate the predicted labels in Attribute-based model with CNN features as a combined representation.
\vspace{-1mm}
\item {\bf{KB\,-\,Affordance (Attributes)}} A smaller KB learned without affordances (attributes).
\vspace{-1mm}
\item {\bf{KB\,-\,Full}} Our full KB model defined in Sec.~\ref{sec:datasource}.
\vspace{-1mm}
\end{itemize}

The Attributes\,+\,Features model (the third row in Table~\ref{table:scene-recogniton}) outperforms the Attribute-based model (the second row in Table~\ref{table:scene-recogniton}) by 11.7\%, indicating the importance of modeling continuous features in the KB.
The full model KB\,-\,Full achieves the state-of-the-art performance on both basic-level and fine-grained classes with more than 2\% improvement over the CNN baseline.

\begin{figure*}
\centering
\includegraphics[width=1.\linewidth]{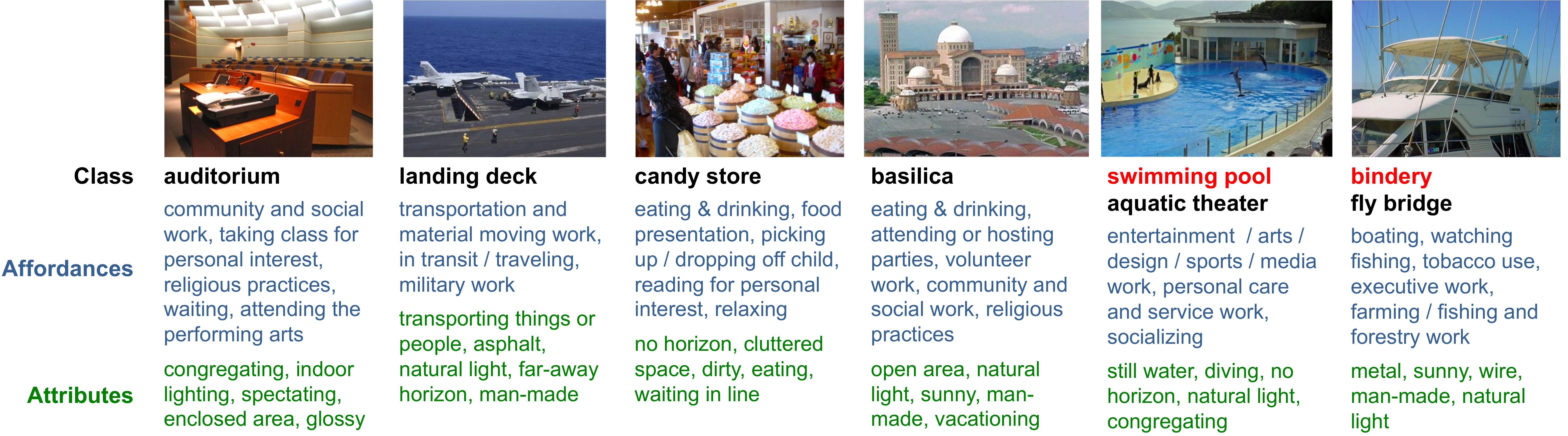}
\vspace{-7mm}
\caption{\small{\textbf{Sample prediction results by the full KB model.} The ground-truth categories (in black) are shown in the first row. The first four images show examples of correct predictions from our KB model, and the last two show incorrect examples. As our model jointly infers multiple labels of an image, we show the predicted affordances (second row) in blue, and the predicted attributes (third row) in green.}}
\label{fig:test-images}
\vspace{-4mm}
\end{figure*}

Fig.~\ref{fig:test-images} offers some insight as to why a KB-based model performs well in a scene classification task. The class label is one of the many labels jointly inferred and predicted by the KB system, including attributes and affordances. So to predict an \emph{auditorium}, attributes such as \emph{indoor lighting}, \emph{enclosed area}, and affordances such as \emph{taking class for personal interest} can all help to reassure the prediction of an auditorium, and vice versa.

As mentioned in Sec.~\ref{sec:datasource}, we have collected annotations of 227 affordance classes for each of the 298 scene categories. We report the performance of {\em affordance prediction} by mean average precision (mAP) and mean F1 score (mF1) over the 227 affordance classes.
The results are presented in Table~\ref{table:affordance_prediction}. Here we compare our full KB model with the CNN Fine-tuned model~\cite{zeiler2014eccv}, where we trained an $\ell_2$-logistic regression classifier on the CNN features for each of the 227 affordance classes. The KB\,-\,Full model outperforms the CNN baselines on both metrics.

\begin{table}[t]
\captionsetup{justification=centering}
\caption{\textbf{Performance of Scene Affordance Prediction}}
\centering
\vspace{-2mm}
\begin{tabular}{|l|c|c|}
\hline
\textbf{Method} & \textbf{mF1} & \textbf{mAP}\\
\hline
\hline
CNN Fine-tuned \cite{zeiler2014eccv} & 81.6 & 74.2\\
KB - Full & \textbf{82.6} & \textbf{75.7}\\
\hline
\end{tabular}
\label{table:affordance_prediction}
\vspace{-2mm}
\end{table}

\begin{figure}[htb]
\centering
\includegraphics[width=1.0\linewidth]{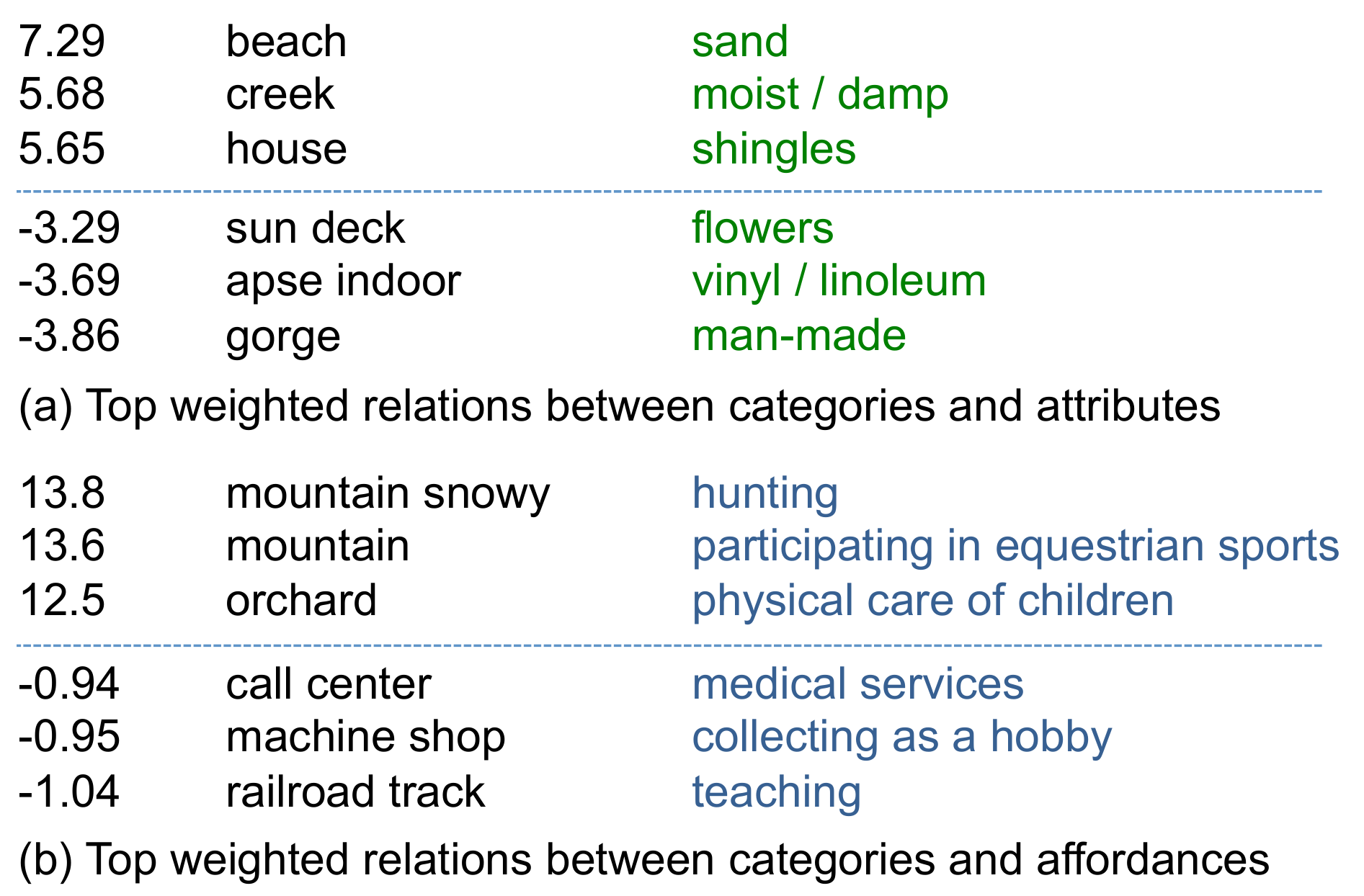}
\vspace{-7mm}
\caption{\small{\textbf{Examples of the strongest and the weakest relations in the learned KB.}
(a) Relations between scene classes (left column) and scene attributes (right column). (b) Relations between scene classes (left column) and scene affordances (right column). In both (a) and (b), the number at the beginning of each row indicates the actual factor weight in the underlying MRF. The more positive the number, the stronger the correlation. We show relations with the largest positive and negative weights in the KB.
To be consistent with Fig.~\ref{fig:test-images}, we use the same color scheme for attributes and affordances.}}
\label{fig:top-rules}
\vspace{-3mm}
\end{figure}

Recall that the KB framework learns the weights of the relations between entities (e.g., scene classes, attributes and affordance, etc.) in a joint fashion. We can then examine the strength of these relations by looking at the factor weights of the underlying MRF. A large positive weight between two entities indicate a strong co-occurrence relation; whereas a large negative weight indicates a strong negative correlation. 
Fig.~\ref{fig:top-rules} provides examples of both the strongest and the weakest correlations between scene classes and attributes (Fig.~\ref{fig:top-rules}(a)), as well as scene classes and affordances (Fig.~\ref{fig:top-rules}(b)). For example, the KB has learned that the class \emph{beach} has a strong co-occurrence relation with the attribute \emph{sand}, and the class \emph{railroad track} lacks correlation with the affordance \emph{teaching}.

\subsection{Image Search by Text Queries}
\label{sec:image-search-by-text}
Using the same model and framework, we can also query our KB for sets of images, instead of just one (Sec.~\ref{sec:scene-recognition-annotation}), such as ``\emph{find me images of a sunny beach}.'' Here we use the same dataset as in Sec.~\ref{sec:scene-recognition-annotation}. This task can also be expressed by a single query where the image is taken as variables (see the example in Sec.~\ref{sec:qa-setup}).

\begin{figure}
\subfigure[]{
\includegraphics[width=.45\linewidth]{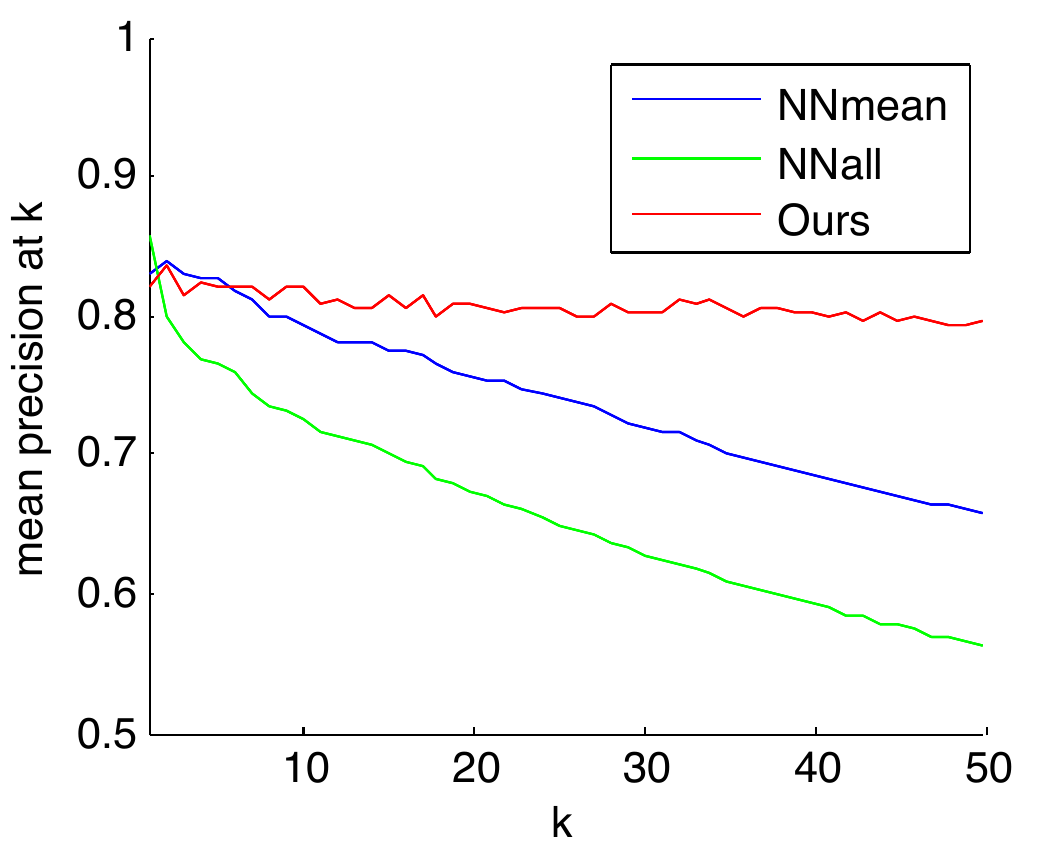}
}
\subfigure[]{
\includegraphics[width=.45\linewidth]{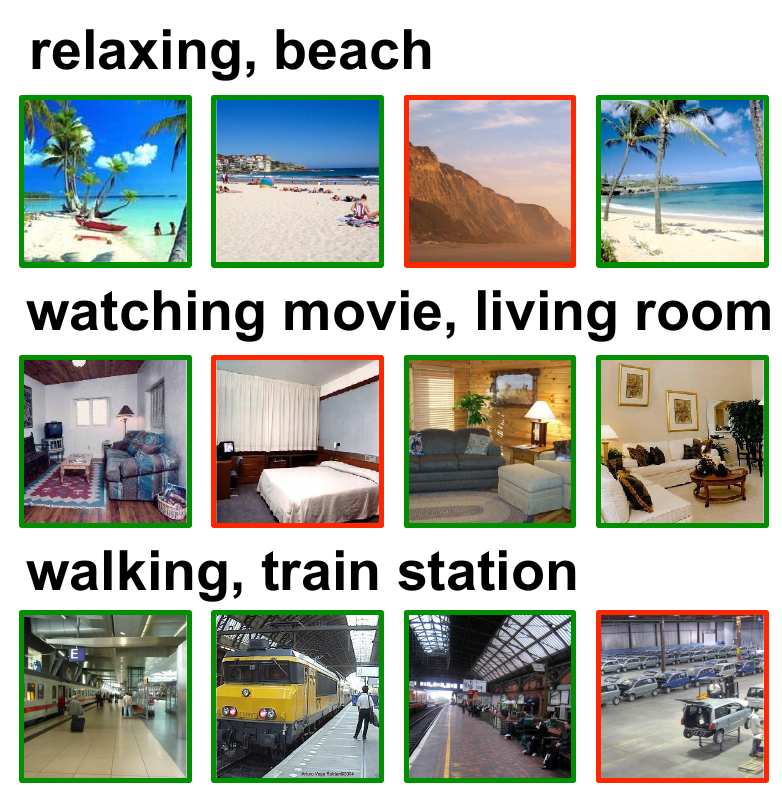}
}
\vspace{-4mm}
\caption{\small{(a) \textbf{Performance variations of top $k$ retrievals} We compare our method with two nearest neighbor baselines. In contrast to these two methods, the KB model maintains a steady performance on lower-ranked retrievals. (b) \textbf{Top retrievals of example queries.} We show top four retrievals from three sample queries (in bold) by our KB model. The green boxes indicate correct retrievals, and red ones indicate incorrect retrievals.}}
\label{fig:retrieval}
\vspace{-4mm}
\end{figure}

We randomly generate 100 queries of a single label (scene category, affordance or attribute), and 100 queries of a pair of labels, each having at least 50 positive samples in the test set. 
Given a set of query labels, we aimed to retrieve the test images that are annotated with all the semantic labels in the set. We compare with two nearest neighbor baseline methods~\cite{heller2006cvpr}. 
NNall ranks the test images based on the minimum Euclidean distance to any individual positive sample in the training set. NNmean ranks the images based on the distance to the centroids of the features of the positive samples. We report  the mean precision at $k$, the mean fraction of correct retrievals out of the top $k$ over all queries, where $k$ goes from 1 to 50.
As shown in Fig.~\ref{fig:retrieval}, our method outperforms both simple nearest neighbor baselines when $k>5$. NNmean performs better than ours among the top five retrievals; however, the false positive rate grows as the number of retrievals increases. In contrast, the relations in the KB compensate the weak and noisy visual signals, and, as a result, maintain stable and good performance on lower-ranked retrievals.

\vspace{-1mm}
\section{Conclusion}
\vspace{-1mm}
This paper presents a principled framework to perform learning and inference on a large-scale multimodal knowledge base (KB). Our contribution is to build a scalable KB to answer a variety of visual queries without re-training. 
Our KB is capable of making predictions on a number of standard vision tasks, on par with state-of-the-art models trained specifically for those tasks. 
In addition to these custom-trained classifiers, it is also interesting to explore these knowledge representations as an attempt towards tackling complex queries in real-world vision applications. Furthermore,  this platform can be used to explore image-based reasoning. Towards these goals, future directions include a tighter integration between language and vision, and a more robust model for incorporating richer information.

{\small
\bibliographystyle{ieee}
\bibliography{refs}
}

\clearpage
\newpage
\appendix

\section{Scalable Knowledge Base Construction}
\label{sec:kbc-supp}
There are three key steps to make the knowledge base construction (KBC) scalable: data pre-processing, factor graph generation and high-performance learning. Sec.~\ref{sec:scalability} provides an overview of the KBC process illustrating these three steps. Here we provide more detailed explanations of our knowledge base construction pipleline.

\subsection{Database Schema}
\label{sec:db-schema}
\begin{figure}[htb]
\centering
\includegraphics[width=.9\linewidth]{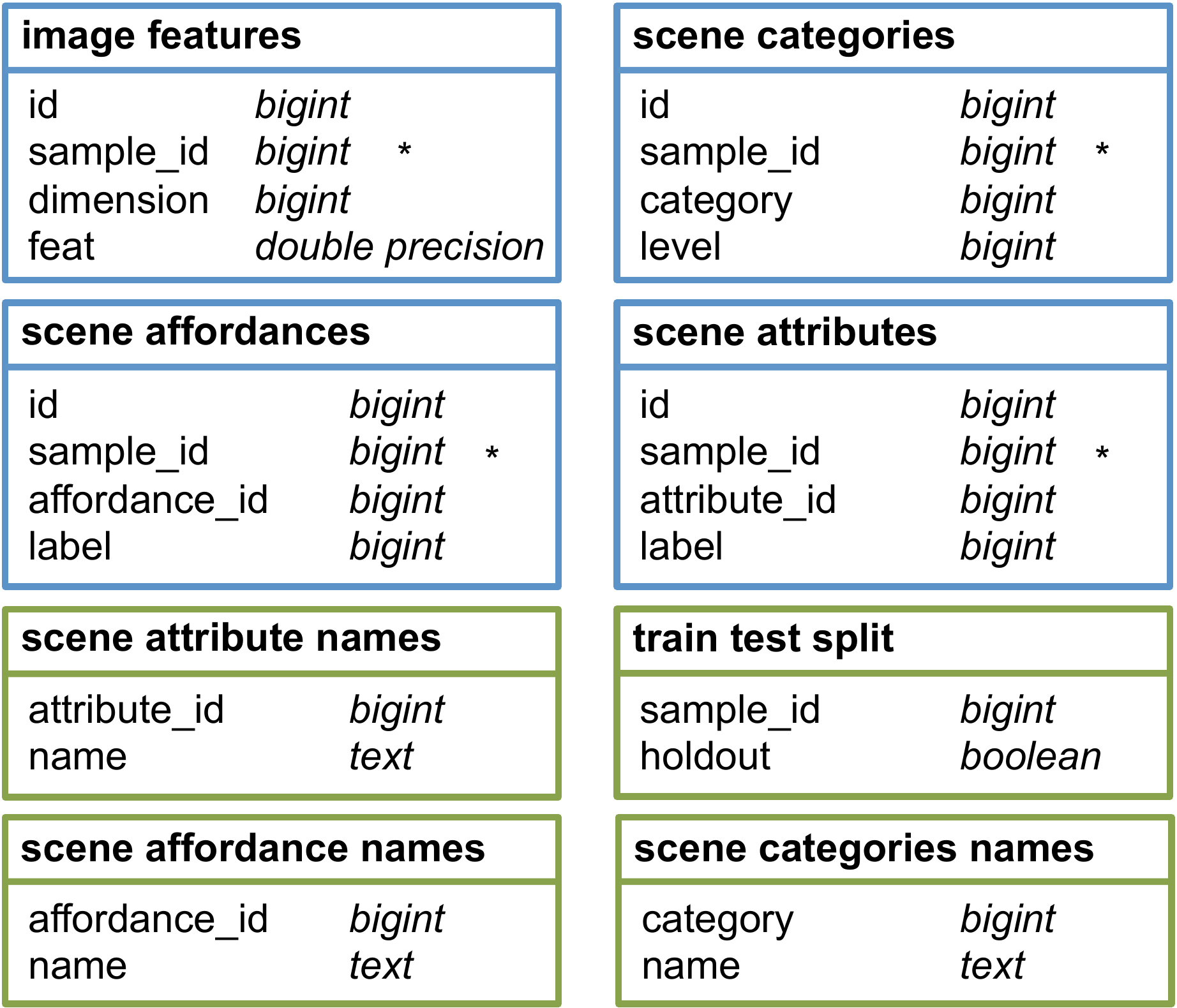}
\caption{\small{\textbf{Database schema for structured representation.} 
The table names (in bold), column names (left) and data types (right) are provided. The blue boxes denote data tables containing KB entities; and the green ones denote metadata tables. The \emph{id} column is a unique identifier for each row, which is used to create the factor graph. The stars (*) indicate the distribution keys for parallel data processing.}}
\label{fig:schema}
\end{figure}

The first step (the first box in Fig.~\ref{fig:kb-pipeline}) is to pre-process raw data into a structured representation. This representation enables us to perform structured queries (e.g. SQL) on the data.
We provide the complete database schema in Fig.~\ref{fig:schema}. 
The schema contains two types of tables: \emph{data tables} contain the entities in Sec.~\ref{sec:datasource} that are used to build the knowledge base (KB); \emph{metadata tables} provide auxiliary information for the experiments and visualization.  \emph{sample\_id} in Fig.~\ref{fig:schema} is a unique identifier of each training sample. These identifiers are used as a distribution key in the database system, where the data is distributed across segments as per the distribution keys.

Each data table stores entities of a certain type. We have a separate table for each of the four entity types in Sec.~\ref{sec:datasource}, where continuous values (image features) are stored as \emph{double precision} numbers, and discrete values (scene category, affordance and attribute labels) are stored as \emph{bigint}. 
We have seen in Sec.~\ref{sec:scalability} that each row in the data tables corresponds to a variable in the factor graph. Thus the entities in Sec.~\ref{sec:datasource} can be represented by different types of variables. We use 4096 continuous variables to represent an \emph{Image} entity by its feature extracted from a fine-tuned CNN~\cite{zeiler2014eccv}. We use a multinomial variable to represent a scene category label, and Boolean variables to represent each of the attribute labels and affordance labels.

\subsection{Runtime environment}
The knowledge base construction is conducted on a Non-Uniform Memory Access (NUMA) machine \cite{zhang2014pvldb} with four NUMA nodes. Each has 12 physical cores and 24 logical cores, with Intel Xeon CPU@2.40GHz and 1TB main memory. We choose Greenplum as the underlying database system due to its power in massive parallel data processing.\footnote{\url{http://www.pivotal.io/big-data/pivotal-greenplum-database}}

\subsection{Human-readable Rules}
\label{sec:language}
To define the KB with ease, we develop a declarative language, which serves as a human-readable interface for specifying the KB structure.
The syntax of the declarative language is an extension to first-order logic in order to accommodate continuous variables.
We introduced in Sec.~\ref{sec:datasource} three types of relations. 
We define each type of relations by a group of rules, where
each rule $R_j$ is a set specified
with first-order logic formulas. 

We first explain an example rule. We then describe the general form of the rules later. In Fig.~\ref{fig:kb-pipeline} we have shown that our KBC system creates a factor in the factor graph of image \texttt{I1} from the rule $\texttt{hasAffordance}(\texttt{I1},\,\texttt{travel})\,\wedge\,\texttt{hasAttribute}(\texttt{I1},\,\texttt{sunny})$, which describes the co-occurrence between the affordance label \texttt{travel} and the attribute label \texttt{sunny}. We use the same example to show how factors are generated from the declarative language. Instead of writing rules for each of the affordance-attribute pair, we can simply write a rule:
\begin{equation*}
\begin{small}
\{(i, w(x,y), 1)\,|\,\text{hasAffordance}(i,x)\,\wedge\,\text{hasAttribute}(i, y)\}
\end{small}
\end{equation*}
where $i$, $x$ and $y$ correspond to the variables of images, affordance labels and attribute labels respectively. This rule can be instantiated by assigning values to these variables. One possible assignment is to set $i$ to image \texttt{I1}, $x$ to \texttt{travel} and $y$ to \texttt{sunny}. This creates a factor in the factor graph of image \texttt{I1}, where the factor value is 1 when $\texttt{hasAffordance}(\texttt{I1},\,\texttt{travel})\,\wedge\,\texttt{hasAttribute}(\texttt{I1},\,\texttt{sunny})$ holds and 0 otherwise. It evaluates to 0 in the example of Fig.~\ref{fig:kb-pipeline}, as image \texttt{I1} does not have attribute \texttt{sunny}. Under such variable assignment, the weight assigned to the factor is $w(\texttt{travel},\texttt{sunny})$. It indicates that this weight will be shared by all the factors (one for each training image) that depict the co-occurrence between the affordance \texttt{travel} and the attribute \texttt{sunny}.
This rule indicates that image
$\texttt{I1}$ should have both
\texttt{hasAffordance($\texttt{I1}$,$\texttt{travel}$)} and 
\texttt{hasAttribute($\texttt{I1}$,$\texttt{sunny}$)}
to be true with a confidence score of
$w(\texttt{travel}, \texttt{sunny})$.
Similarly, the corresponding factors for other images share the same weight $w(\texttt{travel}, \texttt{sunny})$.
More generally, each rule $R_j$ corresponds to a set 
in a given possible world $I$:
\begin{equation}
I(R_j) = \left\{ (\bar{x}, w(\bar{y}), f(\bar{z})) \right\}
\end{equation}
where 
$\bar{x}, \bar{y}, \bar{z}$ are sets of 
variable in the domain (the set of all possible values the variables can take),
and $w(\cdot)$ and $f(\cdot)$ are real-valued functions.
Here $f(\cdot)$ essentially defines factors in the factor graph model and $w(\cdot)$ 
defines the corresponding factor weights (see Sec.~\ref{sec:representation}).
The arguments to $f(\cdot)$ define the variables required to compute the factor value. The arguments to $w(\cdot)$ define how the factor weights are shared across the factors.

All three types of relations in Sec.~\ref{sec:datasource} can be specified as rules written in this declarative language. Fig.~\ref{fig:full_inference_rules} provides a complete list of rules that we have used to build the visual KB. 
To be more specific, we express \emph{image\,-\,label} relations using two sets of rules corresponding to 1) the linear terms, where the factors return the image feature values of each dimension; and 2) the bias terms, where the factors return a constant 1. 
For intra- and inter-correlations, we express them as conjunctions of two  predicates, where the factors return 1 if both labels take the same Boolean value (either true or false), and 0 otherwise.
In total, the proposed declarative language enables us to define the KB structure with eighteen first-order logic rules.
Our KBC system automatically parses these rules, and creates a factor graph (see the second box in Fig.~\ref{fig:kb-pipeline}).
Now we have the structure of the factor graph model, the next step is to learn the model parameters (i.e., factor weights).
We will talk about the details of learning and inference in the next section.

\subsection{Learning and Inference}
\label{sec:learning_and_inference}
In this section we provide more technical details about learning and inference in our KB.

\subsubsection{Learning}

The factor graph model in Sec.~\ref{sec:representation} is an instance of standard energy-based probabilisitic models~\cite{koller2009probabilistic} where the energy function $E(I)$ is defined through a linear combination of factors:
\begin{equation}
E(I)=\sum_{i=1}^mw_if_i(I)
\end{equation}
A standard approach to learning is to optimize the negative log-likelihood of the training data in Eq.~\eqref{eq:learning_objective}. 
Due to the intractability of computing the analytical gradients, sampling is a common practice to estimate the log-likelihood gradients. The gradient approximation used in Eq.~\eqref{eq:approximate-gradient-partial} is a special case of contrastive divergence~\cite{hinton2002training}, called CD-1. Namely, instead of waiting for the Markov chain to converge, we obtain a sample after only one step of Gibbs sampling. This significantly reduces the cost of gradient computation per step, and has shown effective in several learning tasks~\cite{carreira2005aistats,hinton2002training}. We illustrate in Fig.~\ref{fig:kb-pipeline}(d) that we create a factor graph for each image. This process is sometimes called \emph{grounding} in the literature~\cite{Richardson06ml}. During training we treat these small factor graphs as a single large factor graph. The variables are mixed and shuffled before sampling. A weight update is performed at each Gibbs sampling step.

\subsubsection{Inference}
The inference task is to derive the marginal probabilities of a conjunctive query in Eq.~\eqref{eq:answer_prob}. This problem can be regarded as computing the expectation of a real function $f:\mathcal{I}\rightarrow \mathbb{R}$ given the probability distribution of possible worlds $I\in\mathcal{I}$:
\begin{equation}
\mathbf{E}[f;\mathbf{w}] = \sum_{I\in \mathcal{I}}\Pr[I;\mathbf{w}]f(I)
\label{eq:exp-exact}
\end{equation}
where $\Pr[I;\mathbf{w}]$ is the probability of a possible world $I$ defined in Eq.~\eqref{eq:probability_world}, and $\mathcal{I}$ is the set of all possible worlds.
Computing the exact expectation in Eq.~\eqref{eq:exp-exact} is intractable in general factor graphs, which requires summing over a large (or even infinite) number of variable assignments. Gibbs sampling is a commonly used method for approximate inference. 

The Gibbs sampling starts with an initial world $I^{(0)}$. For each random variable $v_k$ in the factor graph, we sample its new value $v_k'$ from the conditional distribution $\Pr[v_k| MB(v_k); \mathbf{w}]$, where $MB(v)$ is the Markov blanket of the variable $v$. In the context of factor graphs~\cite{Kschischang01it}, the Markov blanket of a variable is the set of factors that are connected to the variable. The sampler then moves to the next variable. After $m$ rounds of iterations, we have sampled a collection of possible worlds $\Omega=\{I^{(0)}, I^{(1)}, \ldots, I^{(m)}\}$. We thus approximate the expectations of a query $q$ in Eq.~\eqref{eq:exp-exact} over $\Omega$:
\begin{equation}
\mathbf{\tilde{E}}[q] = \frac{1}{m}\sum_{i=1}^mq(I^{(i)}),
\end{equation}
where $q(I)$ is the value of the conjunctive query $q$ in possible world $I$. To be specific, $q(I)$ evaluates to 1 if all the predicates in the query $q$ are true in the possible world $I$, and 0 otherwise.
After sufficient iterations, the probability of an answer to the query can be estimated by the number of iterations in which it takes that value over the total number of iterations.

\newcommand{\somewhatImportantCode}[1]{\textbf{\textcolor[rgb]{0.0,0.35,0.6}{#1}}}
\newcommand{\veryImportantCode}[1]{\small\textbf{\textsc{{\textcolor[rgb]{0.0,0.5,0.9}{#1}}}}}
\lstset{
  language=C,
  basicstyle=\ttfamily,
  showstringspaces=false,
  breaklines=true,
  keywordstyle={\textit},
  escapechar=\&
}
\begin{figure}[t]
\begin{scriptsize}
\begin{lstlisting}[mathescape]
&\veryImportantCode{Image\,-\,label relations}&

&\somewhatImportantCode{image features \& scene category }&
{(i,w(d),f) | sceneCategory(i,c)$\,\wedge\,$hasFeature(i,d,f)}
{(i,w(c),1) | sceneCategory(i,c)}

&\somewhatImportantCode{image features \& scene affordance}&
scene_affordance_and_scene_features
{(i,w(a),f) | HasAffordance(i,a)$\,\wedge\,$hasFeature(i,d,f)}
{(i,w(a),1) | hasAffordance(i,a)}

&\somewhatImportantCode{image features \& scene attribute}&
{(i,w(d),f) | hasAttribute(i,a)$\,\wedge\,$hasFeature(i,d,f)}
{(i,w(a),1) | hasAttribute(i,a)}

&\veryImportantCode{Intra-correlations}&

&\somewhatImportantCode{affordance \& affordance}&
{((i,a1,a2), w(a1,a2), 1) | hasAffordance(i,a1)$\,\wedge\,$hasAffordance(i,a2)}
{((i,a1,a2), w(a1,a2), 1) | !hasAffordance(i, a1)$\,\wedge\,$
    $\,\,$!hasAffordance(i, a2)}

&\somewhatImportantCode{attribute \& attribute}&
{((i,a1,a2),w(a1,a2),1) | hasAttribute(i,a1)$\,\wedge\,$hasAttribute(i,a2)}
{((i,a1,a2),w(a1,a2),1) | !hasAttribute(i,a1)$\,\wedge\,$
    $\,\,$!hasAttribute(i,a2)}

&\veryImportantCode{Inter-correlations}&

&\somewhatImportantCode{category \& attribute}&
{((i,c,a), w(a,c), 1) | sceneCategory(i, c)$\,\wedge\,$hasAttribute(i, a)}
{((i,c,a), w(a,c), 1) | sceneCategory(i, c)$\,\wedge\,$
    $\,\,$!hasAttribute(i, a)}
{((i,c,a), w(a,c), 1) | !sceneCategory(i, c)$\,\wedge\,$hasAttribute(i, a)}
{((i,c,a), w(a,c), 1) | !sceneCategory(i, c)$\,\wedge\,$
    $\,\,$!hasAttribute(i, a)}

&\somewhatImportantCode{category \& affordance}&
{((i,c,a), w(a,c), 1) | sceneCategory(i, c)$\,\wedge\,$hasAffordance(i, a)}
{((i,c,a), w(a,c), 1) | sceneCategory(i, c)$\,\wedge\,$
    $\,\,$!hasAffordance(i, a)}
{((i,c,a), w(a,c), 1) | !sceneCategory(i, c)$\,\wedge\,$hasAffordance(i, a)}
{((i,c,a), w(a,c), 1) | !sceneCategory(i, c)$\,\wedge\,$
    $\,\,$!hasAffordance(i, a)}

\end{lstlisting}
\end{scriptsize}
\caption{\textbf{The complete list of rules for the visual knowledge base construction.} We build our visual knowledge base with the rules above. $!$ denotes negation and $\wedge$ denotes conjunction. The formal semantics of the rules are described in Sec.~\ref{sec:language}.}
\label{fig:full_inference_rules}
\end{figure}

\section{Query Answering Application Setup}
\label{sec:qa-app-supp}
In Fig.~\ref{fig:crazy-queries}, we have provided six query examples that illustrate the diversity of tasks our KB system can handle. In order to answer these diverse types of queries, it requires a fusion of information from various sources. In practice, we aggregate information from online databases, business and travel websites, etc. We provide the detailed experimental setups and the data sources here.

We augment our KB in Sec.~\ref{sec:datasource} with a new set of geo-tagged images and several types of metadata. We briefly introduce the extra data sources that we used for
this experiment in Sec.~\ref{sec:exp-qa}.
We randomly sample from Flickr100M\footnote{\url{http://yahoolabs.tumblr.com/post/89783581601/one-hundred-million-creative-commons-flickr-images}} a pool of 20k images with geo-tags and timestamps. 
Besides these images, we incorporate additional information by either downloading from existing databases or crawling from the web.
All the information is stored in a structured format as database tables (Sec.~\ref{sec:db-schema}). 

\begin{enumerate}
\item We obtain a list of names and dates of 327 public holidays  from
Freebase\footnote{\url{https://www.freebase.com}}~\cite{bollacker2008sigmod} from the instances of \texttt{/time/holiday\_category/holidays}.
\item We scrape business information from Yelp.com and Hotels.com. We have crawled in total over sixteen thousand entries of business information, including 7k bars, 6k shopping centers and 3k hotels.
\item We download the daily temperature and weather data from 
National Climatic Data Center. Climate Data Online\footnote{\url{http://www.ncdc.noaa.gov/cdo-web/}} (CDO) provides free access to global historical weather and climate data.
\item We download the publicly available GeoNames geographical database\footnote{\url{http://www.geonames.org/}}, which maps geolocations to over eight million place names.
\end{enumerate}

We introduce new predicates in Fig.~\ref{fig:example_queries} (Boolean-valued functions) that
enable us to query with these additional data. The semantics of these new predicates  can be easily inferred from the predicate names and input variables. For instance, the predicate $\texttt{hasLocation}(img, \texttt{latlong1})$ evaluates to true if the image $img$ was annotated with the geo-location \texttt{latlong1} and false otherwise; $\texttt{nearBy}(latlong1, latlong2, \texttt{1km})$ evaluates to true if the two geo-locations are within 1km away and false otherwise.
Having defined the predicates, we use the augmented KB to answer the queries in Fig.~\ref{fig:crazy-queries}. 
We list the conjunctive queries for each of the six example queries in Fig.~\ref{fig:example_queries}.
The predicates in each query are connected by logical conjunctions. Therefore the query evaluates to 1 if and only if every predicate in the query is true, and 0 otherwise.
\texttt{answer($\cdot$)} indicates the return variables, i.e., the target answers to the queries. We retrieve a ranked list of the answers by computing a marginal probability of the queries (see Sec.~\ref{sec:qa-setup} and Sec.~\ref{sec:learning_and_inference}). Note that, once these additional metadata are incorporated into the KB framework, our system treats images, existing metadata and these new metadata on an equal footing in learning and inference. Therefore, a query can be answered by a joint inference with no post-filtering steps.

\begin{figure}
$\begin{small}
\textbf{Q: Find me a modern looking mall near Fisherman's Wharf.}\\
\text{hasLocation}(img, latlong1)\\
\text{mall}(mall, latlong2, zip)\\
\text{geoName}(\texttt{Fisherman's Wharf}, latlong3)\\
\text{hasAttribute}(img, \texttt{indoor lighting})\\
\text{hasAttribute}(img, \texttt{glossy})\\
\text{nearBy}(latlong1, latlong2, \texttt{1km})\\
\text{nearBy}(latlong1, latlong3, \texttt{20km})\\
\Rightarrow \text{answer}(img, mall, zip)\\
\vspace{-2mm}
\\
\textbf{Q: Find me a place in Boston where I can play baseball.}\\
\text{hasAffordance}(img, \texttt{playing baseball})\\
\text{hasLocation}(img, latlong1)\\
\text{geoName}(\texttt{Boston}, latlong2)\\
\text{nearBy}(latlong1, latlong2, \texttt{1km})\\
\Rightarrow \text{answer}(img, latlong1)\\
\vspace{-2mm}
\\
\textbf{Q: Find me a hotel in Boston with new furniture.}\\
\text{hasLocation}(img, latlong1)\\
\text{hasAttribute}(img, \texttt{glossy})\\
\text{geoName}(\texttt{Boston}, latlong2)\\
\text{nearBy}(latlong1, latlong2, \texttt{20km})\\
\text{hotel}(hotel, latlong2, date, price, phone)\\
\Rightarrow \text{answer}(img, hotel, price, phone)\\
\vspace{-2mm}
\\
\textbf{Q: Find me a cozy bar to drink beer near the AT\&T Plaza.}\\
\text{hasAttribute}(img, \texttt{cluttered space})\\
\text{hasLocation}(img, latlong1)\\
\text{bar}(bar, latlong2, price, phone)\\
\text{geoName}(\texttt{AT\&T Plaza}, latlong3)\\
\text{nearBy}(latlong1, latlong2, \texttt{1km})\\
\text{nearBy}(latlong1, latlong3, \texttt{1km})\\
\Rightarrow \text{answer}(img, bar, price, phone)\\
\vspace{-2mm}
\\
\textbf{Q: Find me a sunny and warm beach during Christmas Day 2013.}\\
\text{sceneCategory}(img, \texttt{beach})\\
\text{hasAttribute}(img, \texttt{sunny})\\
\text{hasAttribute}(img, \texttt{warm})\\
\text{hasLocation}(img, latlong1)\\
\text{geoName}(location, latlong2)\\
\text{nearBy}(latlong1, latlong2, \texttt{1km})\\
\text{temperature}(location, degree, \texttt{2013/12/25}) \\
\Rightarrow \text{answer}(img, location, degree, latlong2)\\
\vspace{-2mm}
\\
\textbf{Q: Find me pictures of sunny days of
Seattle during August.}\\
\text{hasAttribute}(img, \texttt{sunny})\\
\text{hasLocation}(img, latlong1)\\
\text{hasDate}(img, day, \texttt{August}, year)\\
\text{geoName}(\texttt{Seattle}, latlong2)\\
\text{nearBy}(latlong1, latlong2, \texttt{20km})\\
\Rightarrow \text{answer}(img, day, \texttt{August}, year)
\end{small}$
\vspace{-2mm}
\caption{Conjunctive queries for the query answering examples in Fig.~\ref{fig:crazy-queries}. We omit the conjunction symbols ($\wedge$) between predicates for neatness.}
\label{fig:example_queries}
\vspace{-2mm}
\end{figure}

Following this approach, we are able to express richer and more complex queries by joining different pieces of information with logical conjunctions. 
As we can see, the query language in Sec.~\ref{sec:qa-setup} is capable of expressing a wide range of queries. Moreover, these queries can be answered in a principled manner, by evaluating marginals in the joint probability model. Given such a flexible framework, data becomes the key to extend our model's power of answering real-world questions. We are interested in exploring more efficient and automatic ways to aggregate information from large-scale multimodal corpora for future work.

\section{Affordance Annotations}
\label{sec:affordance-list}
\begin{figure}
\includegraphics[width=1.0\linewidth]{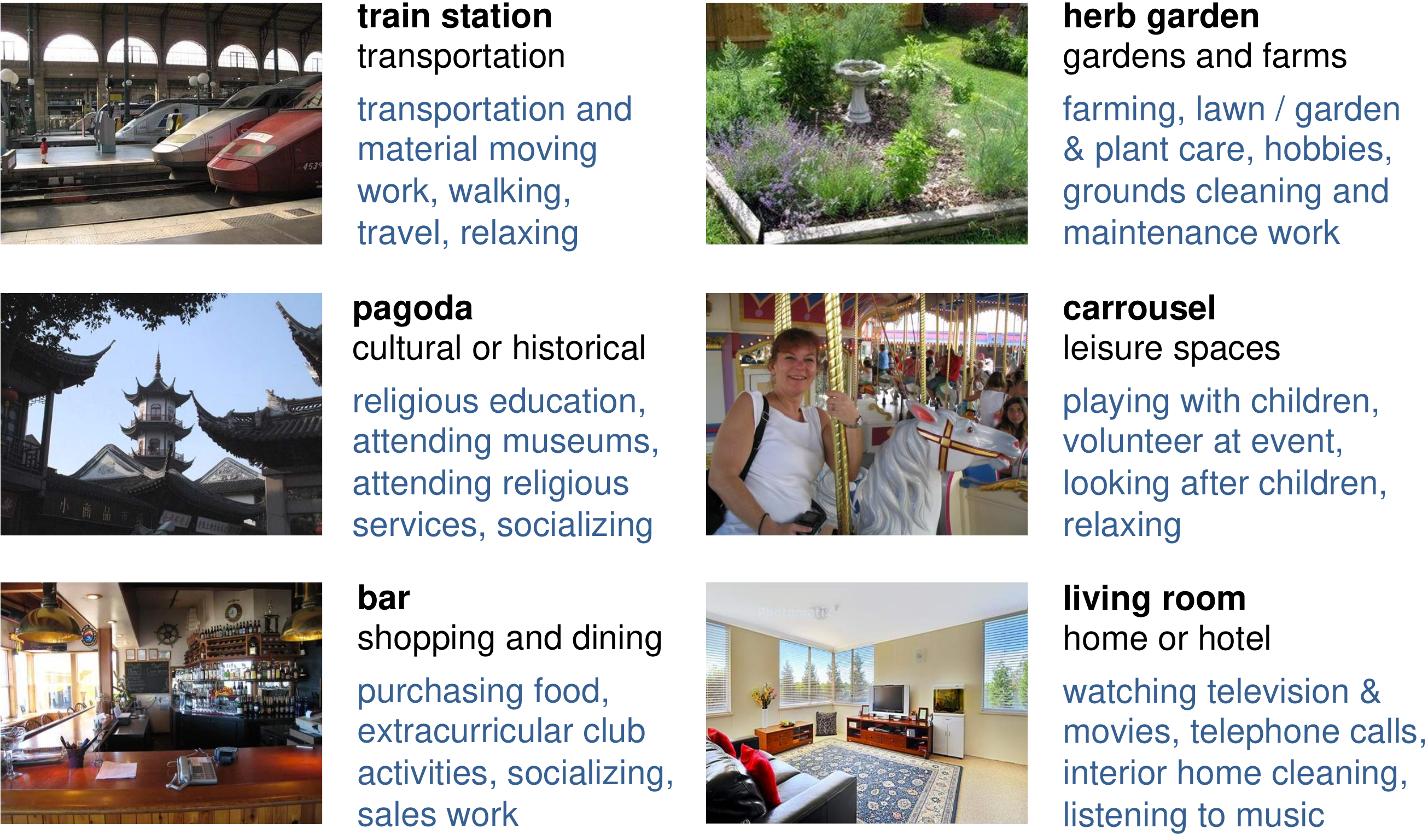}
\caption{\small{\textbf{Sample affordance annotations in the augmented scene dataset.} We augment the SUN dataset~\cite{xiao2010cvpr} with a lexicon of 227 affordances.
We provide the fine-grained category (in bold), the basic-level category and a subset of their affordance annotations.}}
\label{fig:kb-random-samples}
\end{figure}

We augment the SUN dataset~\cite{xiao2010cvpr} with additional annotations of scene affordances. We use a lexicon of 227 affordances (actions) from the American Time Use Survey (ATUS) \cite{shelley2005mlr} sponsored by the Bureau of Labor Statistics, which catalogs the actions in daily lives and represents United States census data. The original ATUS lexicon includes 428 specific activities organized into 17 major activity categories and 105 mid-level categories. We re-organize the categories by collapsing visually similar superordinate categories into one action. For instance, the superordinate-level category ``traveling'' was collapsed into a single category because being in transit to go to school should be visually indistinguishable from being in transit to go to the doctor. This results in 227 actions in total. Fig.~\ref{fig:kb-random-samples} shows six example images with a subset of their affordance annotations. 

The lexicon covers a broad space of possible actions that could take place in scenes. We conducted a large-scale online experiment with over 400 AMT workers annotating the possibilities of the 227 actions for each of the 298 scene categories (Sec.~\ref{sec:datasource}). 10 votes are collected for each category-affordance pair. Positive ($\geq3$ votes) and negative ($\leq 2$ votes) annotations are selected as evidence. 
These 227 affordances are listed in alphabetic order below:

\begin{small}
\paragraph*{A}{appliance repair \& maintenance (self), architecture and engineering work, arts \& crafts, arts \& crafts with children, arts / design / entertainment / sports / media work, attending child's events, attending meetings for personal interest, attending movies, attending museums, attending or hosting parties, attending religious services, attending school-related meetings \& conferences, attending the performing arts}

\vspace{-2mm}
\paragraph*{B}{banking, biking, boating, bowling, building \& repairing furniture, building and grounds cleaning and maintenance work, business and financial operations work, buying / selling real estate}

\vspace{-2mm}
\paragraph*{C}{camping, civic obligations, cleaning home exterior, collecting as a hobby, community and social work, comparison shopping, computer and mathematical work, computer use (not games), construction and extraction work}

\vspace{-2mm}
\paragraph*{D}{dancing, doing aerobics, doing gymnastics, doing martial arts}

\vspace{-5mm}
\paragraph*{E}{eating \& drinking, education and library work, education-related administrative activities, email, exercising \& playing with animals, exterior home repair \& decoration, extracurricular club activities}

\vspace{-2mm}
\paragraph*{F}{farming / fishing and forestry work, fencing, financial management, fishing, food \& drink preparation, food preparation and serving work, food presentation}

\vspace{-2mm}
\paragraph*{G}{gambling, golfing, grocery shopping}

\vspace{-2mm}
\paragraph*{H}{health-related self care, healthcare work, helping adult, helping child with homework, hiking, hobbies, home heating / cooling, home security, home-schooling children, homework, household organization \& planning, hunting}

\vspace{-2mm}
\paragraph*{I}{in transit / traveling, income-generating hobbies \& crafts, income-generating performance, income-generating rental property activity, income-generating selling activities, income-generating services, installation / maintenance and repair work, interior decoration \& repair, interior home cleaning}

\vspace{-2mm}
\paragraph*{J}{job interviewing, job search activities}

\vspace{-2mm}
\paragraph*{K}{kitchen \& food clean-up}

\vspace{-2mm}
\paragraph*{L}{laundry, lawn / garden \& plant care, legal work, listening to music (not radio), listening to radio, looking after adult, looking after children}

\vspace{-2mm}
\paragraph*{M}{mailing, maintaining home pool / pond / hot tub, management / executive work, military work}

\vspace{-2mm}
\paragraph*{N}{non-veterinary pet care}

\vspace{-2mm}
\paragraph*{O}{obtaining licenses \& paying fees, obtaining medical care for adult, obtaining medical care for child, office and administrative work, organizing \& planning for adults, organizing \& planning for children, out-of-home medical services}

\vspace{-2mm}
\paragraph*{P}{participating in aquatic sports, participating in equestrian sports, participating in rodeo, personal care and service work, physical care of adults, physical care of children, picking up / dropping off adult, picking up / dropping off child, playing baseball, playing basketball, playing billiards, playing football, playing games, playing hockey, playing racquet sports, playing rugby, playing soccer, playing softball, playing sports with children, playing volleyball, playing with children (not sports), production work, protective services work, providing medical care to adult, providing medical care to child, purchasing food (not groceries), purchasing gasoline}

\vspace{-2mm}
\paragraph*{R}{reading for personal interest, reading with children, relaxing, religious education, religious practices, rock climbing / caving, rollerblading / skateboarding, running}

\vspace{-2mm}
\paragraph*{S}{sales work, school music activities, science work, security screening, sewing \& repairing textiles, sexual activity, shopping (except food and gas), skiing / ice skating / snowboarding, sleeping, socializing, storing household items, student government}

\vspace{-2mm}
\paragraph*{T}{taking class for degree or certification, taking class for personal interest, talking with children, telephone calls, tobacco use, transportation and material moving work, travel, using cardiovascular equipment}

\vspace{-2mm}
\paragraph*{U}{using clothing repair \& cleaning services, using home repair \& construction services, using in-home medical services, using interior home cleaning services, using lawn \& garden services, using legal services, using meal preparation services, using other financial services, using paid childcare services, using personal care services, using pet services, using police \& fire services, using professional}
\end{small}

\end{document}